\documentclass[conference]{ieeeconf}
\usepackage{cite}
\usepackage{amsmath,amssymb,amsfonts}
\usepackage{algorithmic}
\usepackage{graphicx}
\usepackage{textcomp}
\usepackage{arydshln}
\usepackage{xcolor}
\usepackage{todonotes}
\usepackage{array}
\usepackage{subcaption} % For subfigures
\usepackage{makecell}
\usepackage{booktabs}  
\usepackage{wasysym} 
\usepackage{placeins} 

\def\BibTeX{{\rm B\kern-.05em{\sc i\kern-.025em b}\kern-.08em
    T\kern-.1667em\lower.7ex\hbox{E}\kern-.125emX}}
\begin{document}

\title{\textbf{Soft Manipulation Surface With Reduced Actuator Density For Heterogeneous Object Manipulation}}
% {\footnotesize 
% \textsuperscript{*}Note: Sub-titles are not captured in Xplore and
% should not be used}
% \thanks{Identify applicable funding agency here. If none, delete this.}
% }
\author{
    Pratik Ingle, Kasper Støy, Andres Faiña\\
    \small IT University, Denmark\\ 
    \small \{prin, ksty, anfv\}@itu.dk
}
\maketitle
\begin{abstract}
Object manipulation in robotics faces challenges due to diverse object shapes, sizes, and fragility. Gripper-based methods offer precision and low degrees of freedom (DOF) but the gripper limits the kind of objects to grasp. On the other hand, surface-based approaches provide flexibility for handling fragile and heterogeneous objects but require numerous actuators, increasing complexity. We propose new manipulation hardware that utilizes equally spaced linear actuators placed vertically and connected by a soft surface. In this setup, object manipulation occurs on the soft surface through coordinated movements of the surrounding actuators. This approach requires fewer actuators to cover a large manipulation area, offering a cost-effective solution with a lower DOF compared to dense actuator arrays. It also effectively handles heterogeneous objects of varying shapes and weights, even when they are significantly smaller than the distance between actuators. This method is particularly suitable for managing highly fragile objects in the food industry.

\footnotetext{Funded by the European Union. Views and opinions expressed are however those of the author(s) only and do not necessarily reflect those of the European Union or the European Commission. Neither the European Union nor the granting authority can be held responsible for them.}

\end{abstract}
{\small \textbf{\textit{Index Terms}} -- Manipulation, Soft Robot, Surface Manipulation, Robotic Manipulation Surface}

\section{Introduction}
\begin{figure}[ht]
    \centering
    \includegraphics[width=\linewidth]{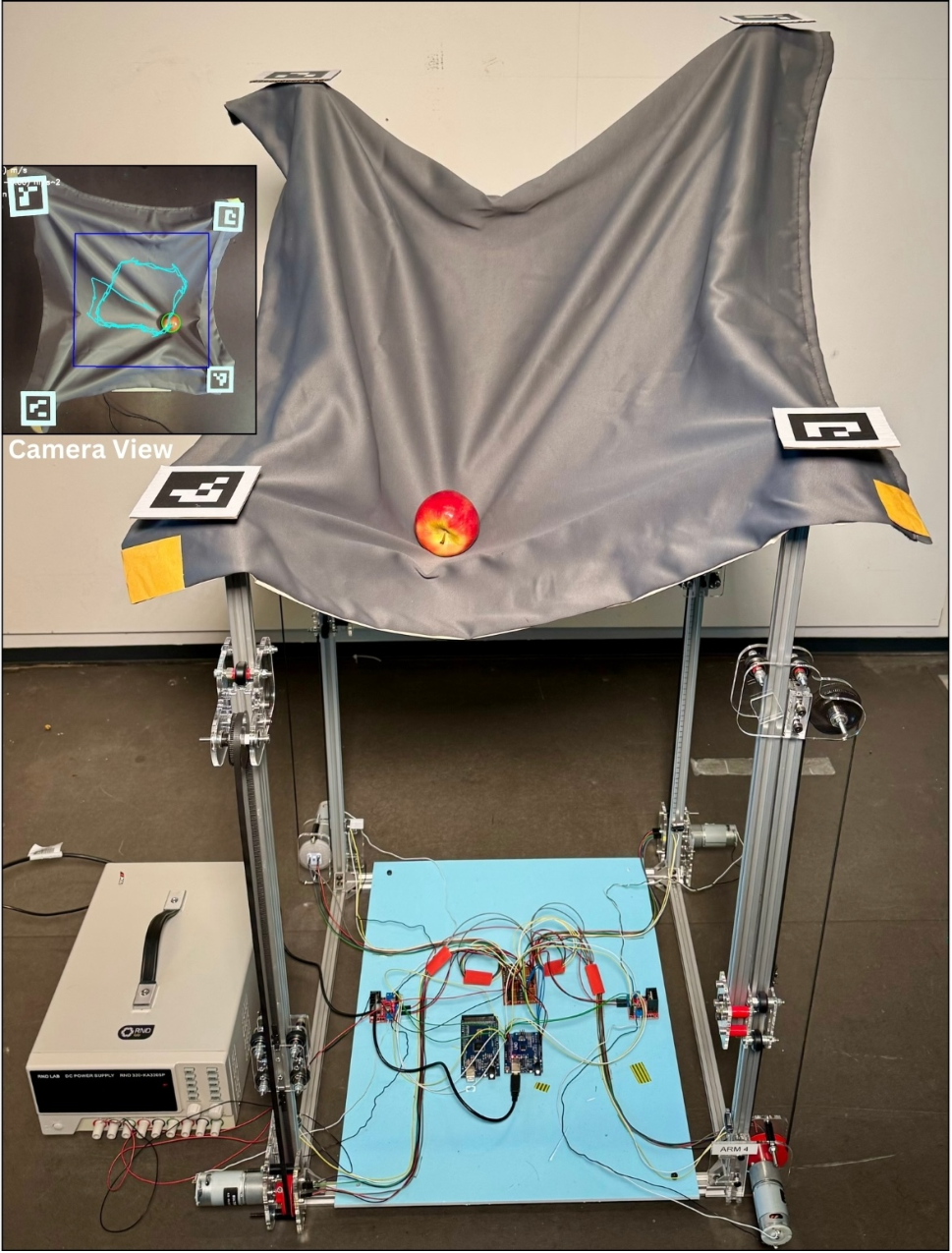}
    \caption{\textbf{Soft Manipulation Surface:} A soft fabric is attached to four actuators positioned at the corners of a $0.5 \times 0.5$-meter aluminium frame. The system guides an apple in a circular path using an Arduino Uno microcontroller. The overhead camera view (top left) highlights the blue region of interest, with the object's trajectory over time shown in light blue, tracked with ArUco markers}
    \label{fig:ModuleSetup}
\end{figure}
Traditional robotic manipulation primarily uses gripper-based methods. A robotic arm equipped with a gripper is employed to pick, push, or pull objects for manipulation. These manipulators are designed with rigid structures, making them versatile and precise, and are well-suited for handling objects with low degrees of freedom (DOF). Grippers offer high precision and control, particularly when integrated with advanced sensors, making them ideal for tasks that require careful handling, such as positioning parts for assembly or packaging items. Despite their strengths, gripper-based systems have certain limitations. They lack the capability for parallel manipulation, meaning they cannot handle multiple objects simultaneously, and they struggle with heterogeneous objects of irregular shapes, soft materials, or those needing delicate force control, as improper pressure can cause damage or lead to dropped items. 

Alternatively, Robotic Manipulation Surfaces (RMS) manipulate objects indirectly by controlling the movement of objects across the surface of a grid of actuators, typically through the use of small actuators or motors that change the surface’s shape or properties. These systems do not grip objects directly but rather rely on the coordinated movement of actuators to push, tilt, or move items smoothly across the surface. The use of multiple actuators provides a large contact area, making them suitable for handling flat items with larger surface areas and reduce the risk of damage. Additionally, RMS can offer flexibility in handling objects of varying sizes and shapes, as multiple actuators can work in parallel, making these systems more efficient for large-scale operations. 

In the literature, several RMS techniques have been explored, primarily using wheel-based or piston-based mechanisms. Wheel-based systems transport objects over long distances, with most manipulation achieved through varying wheel speeds \cite{parajuli2014actuator, uriarte2022methode}. These systems are highly effective for high-throughput operations like sorting and transporting in packaging, distribution, and assembly lines, offering cost-effective movement over long distances and easy integration with other automation systems. However, they are limited to flat and relatively large objects. 

Piston-based manipulation surfaces \cite{xue2024arraybot, johnson2023multifunctional, hashem2016control, follmer2013inform, yu2008morpho} involve using multiple pistons, often smaller than the object, arranged in a dense formation. The relative positioning of the actuators changes the surface beneath the object, enabling movement, rotation, flipping, or sliding. In piston-based RMS, manipulation efficiency decreases—or fails entirely—when the object's size is comparable to either the gap between pistons or the surface area of an individual piston. If the object is smaller than the gap between actuators, it will fall through. Conversely, if the object’s size is close to the surface area of a single actuator, movement will be limited to actions like up-and-down motion. This limitation necessitates densely packed actuator arrays in most surface manipulation systems. Other methods for RMS include vibration-based manipulation \cite{georgilas2015cellular,zhou2016controlling}, air film manipulation \cite{moon2006distributed}, cilia-based surfaces \cite{ataka20052d, yim2000two} or soft gel actuators \cite{tadokoro2000distributed}, each with distinct advantages and limitations. 

Independently of their type, most RMS present a high number of actuators and they can be designed with either centralized or decentralized control and sensing capabilities. However, managing a large number of actuators presents challenges due to high costs and the complexity of controlling multiple DOF.

To address these challenges, we propose a novel approach that incorporates a soft surface layer between equally spaced actuators, figure \ref{fig:ModuleSetup}. In this configuration, most object manipulation occurs on the soft surface through coordinated movements of the surrounding actuators. This reduces the overall number of actuators required to cover large areas, while still maintaining the same level of manipulation capability. By lowering the density of actuators and degrees of freedom, this approach offers a more cost-effective solution compared to traditional manipulation surface systems and can handle objects of heterogeneous objects of various shapes, sizes and weights, including small objects as small as 0.5 cm

In the past, Festo \cite{festo} developed a wave-handling surface robot consisting of a soft, flexible layer covering a large array of pneumatic actuators. This system produced various wave patterns on the soft layer to manipulate objects but was limited to spherical objects and required a dense grid of many actuators. Morpho \cite{yu2008morpho}, a self-deformable modular robot inspired by cellular structures, allows large-scale shape changes and dynamic forms to achieve different global functions. It uses neoprene and latex for the surface membranes and has potential applications for manipulating objects on top of the soft layer, but it does not focus on object manipulation within a single module, only on moving a ball between different modules. Most RMS have an object-to-module size ratio (the smallest object they can handle) greater than or equal to 1, indicating they can only work with objects as large as or larger than a single module. For the Festo system, this ratio is nearly 1. In contrast, our system can handle objects of various sizes, including small objects as small as 0.5 cm, resulting in an object-to-module size ratio of 0.01, and it requires significantly fewer actuators to manipulate them effectively on same size of area.

The use of a soft layer allows the manipulation of highly fragile objects without damaging them. To evaluate the advantages and limitations of this approach, we constructed a hardware setup with a $0.5 \times 0.5$-meter actuator grid, where each actuator can move up to 0.5 m in height. The system successfully manipulated objects of various shapes, sizes, and weights, including fragile items like eggs and fruit (e.g., apples). During the experiments, one of the main challenges identified was the control of object manipulation on the soft surface, due to its non-linear behavior. Despite this, the system demonstrated strong potential for applications where gentle and precise handling of heterogeneous objects is required. To overcome the nonlinear behaviour, we tested a Reinforcement Learning (RL) policy trained in simulation and zero-shot learning to manipulate a cube on the actual hardware.

\section{Methods}
\begin{figure}[ht]
    \centering
    \includegraphics[width=\linewidth]{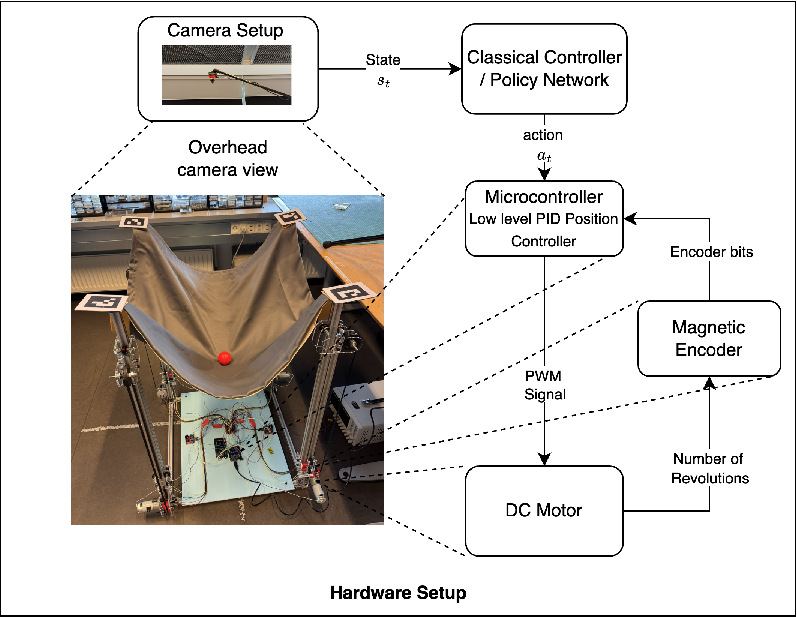} % Adjust the width as needed
    \caption{Hardware setup of the system, including four actuators and an overhead camera for object tracking. The system utilizes an Arduino Uno microcontroller to send PWM signals to the DC motors and to read motor rotations via an AS500 magnetic encoder.}
    \label{fig:hardware_setup}
\end{figure}

In our approach, the actuators are connected using a soft layer, which eliminates the dependency on object size. The manipulation hardware consists of four linear actuators capable of vertical movement, placed at the corners of a square ($0.5 \times 0.5$-meters), each with an effective range of 0.5 meters. Each actuator is powered by a DC geared motor (RS pro 834-7590, 12v, 120 rpm) that converts rotational motion into linear motion using a pulley and belt system. A single actuator with a pulley with a diameter of 5cm and a motor with a max torque of 0.29 Nm gives a force of $\sim$ 11.6 N. The four actuators are mounted vertically on a frame composed of aluminium profiles.

A V-slot system provides a linear guide to the carriage, consisting of two sets of V-slot wheels on top of the linear actuator for smooth linear motion. Each actuator is composed of a vertical V-slot profile (frame) with the motor and pulley clamped at the bottom and a return pulley at the top, secured with acrylic brackets. A carriage, which slides along the vertical V-slot profile using rolling V-slot wheels with bearings and is driven by a belt, is used to move another vertical aluminum profile that acts as a piston. This piston is also guided by an additional set of V-slot wheels clamped at the top of the frame to minimize flexion under external forces. This design provides flexibility for switching between different soft layers and allows for continuous surfaces if more actuators are added in the future.

For precise control, we use an AS500 magnetic encoder with 12-bit resolution to monitor motor rotations. An 8-channel I2C hub addresses the limitation that all four magnetic encoder sensors share the same address. An Arduino Uno controls the system by sending PWM signals to an H-bridge that drives the motors, with a PID controller managing position control to ensure smooth and accurate movements. A $0.6 \times 0.6$-meter, 100\% polyester soft fabric, attached to the corners of the actuators, forms a flexible, hanging surface. Most object manipulation occurs on this soft surface through pulling, rolling and sliding motions driven by the interaction between the object and the fabric. The soft surface provides significant flexibility for handling objects with different textures and levels of smoothness. Depending on the properties of the objects, the surface characteristics can be adjusted for factors such as stretchability or varying levels of friction. 

An overhead camera tracks the object’s position on the fabric surface at a frequency of 12 Hz. ArUco markers placed at each corner of the $0.5 \times 0.5$-meter manipulation region enable accurate real-world tracking within that space. For simpler tracking, we used red-colored objects; however, with more advanced object-tracking techniques, objects of any color could be used. This setup, comprising four actuators, functions as a single module.

\section{Experiments / Results}
Manipulating an object on a soft surface is challenging due to the non-linear behavior of the surface and the complexity of handling it. To test the effectiveness of our system, we conducted two types of experiments on the actual hardware setup:

	1.	Manipulation Dynamics: This experiment was conducted to determine how the relative elevation of the actuators (pistons) affects the dynamic path of the object, depending on its shape. It helps us understand how much slope is required to move an object efficiently across the surface.
 
	2.	Object Behavior: In this experiment, various objects were manipulated to follow a circular trajectory on the fabric. The goal was to map out blind spots on the surface where manipulation becomes difficult, and to observe actions that may cause the object to fall out of the manipulation region within a single module. This information is crucial for scenarios involving multiple modules, especially for transferring objects between modules.
 
The following subsection describe the experimental setup and results of these experiments in more detail. All the experiments focus on the object's position in 2D space (along the xy plane) and do not consider the position along the Z-axis.

\subsection{Manipulation Dynamics}

\begin{figure*}[ht]
    % \centering
    % First row with three columns
    \begin{subfigure}[b]{0.32\textwidth}
        \centering
        \includegraphics[width=\textwidth]{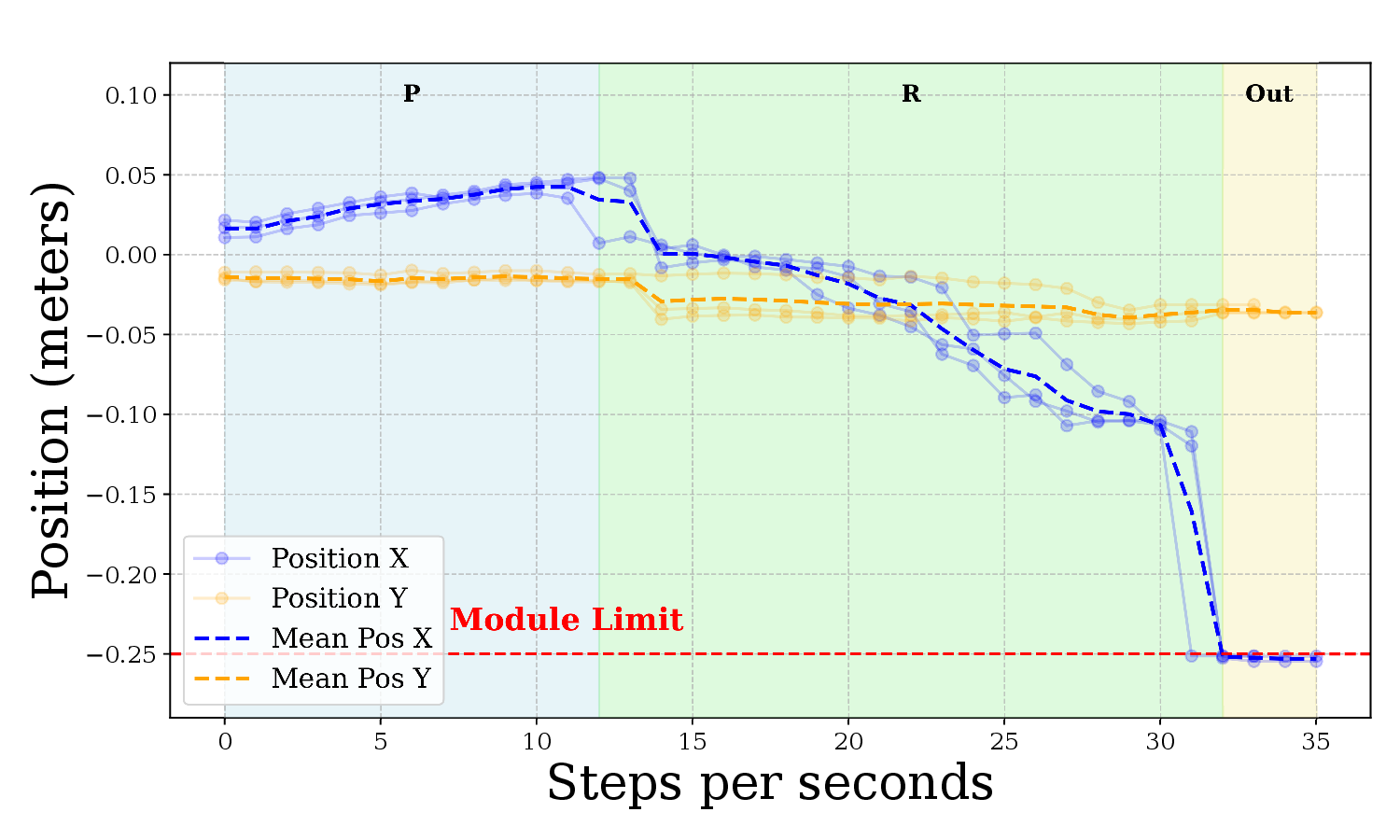} 
        \caption{Sphere dynamics}
        \label{fig:r_edge_sphere}
    \end{subfigure}
    \hfill
    \begin{subfigure}[b]{0.32\textwidth}
        \centering
        \includegraphics[width=\textwidth]{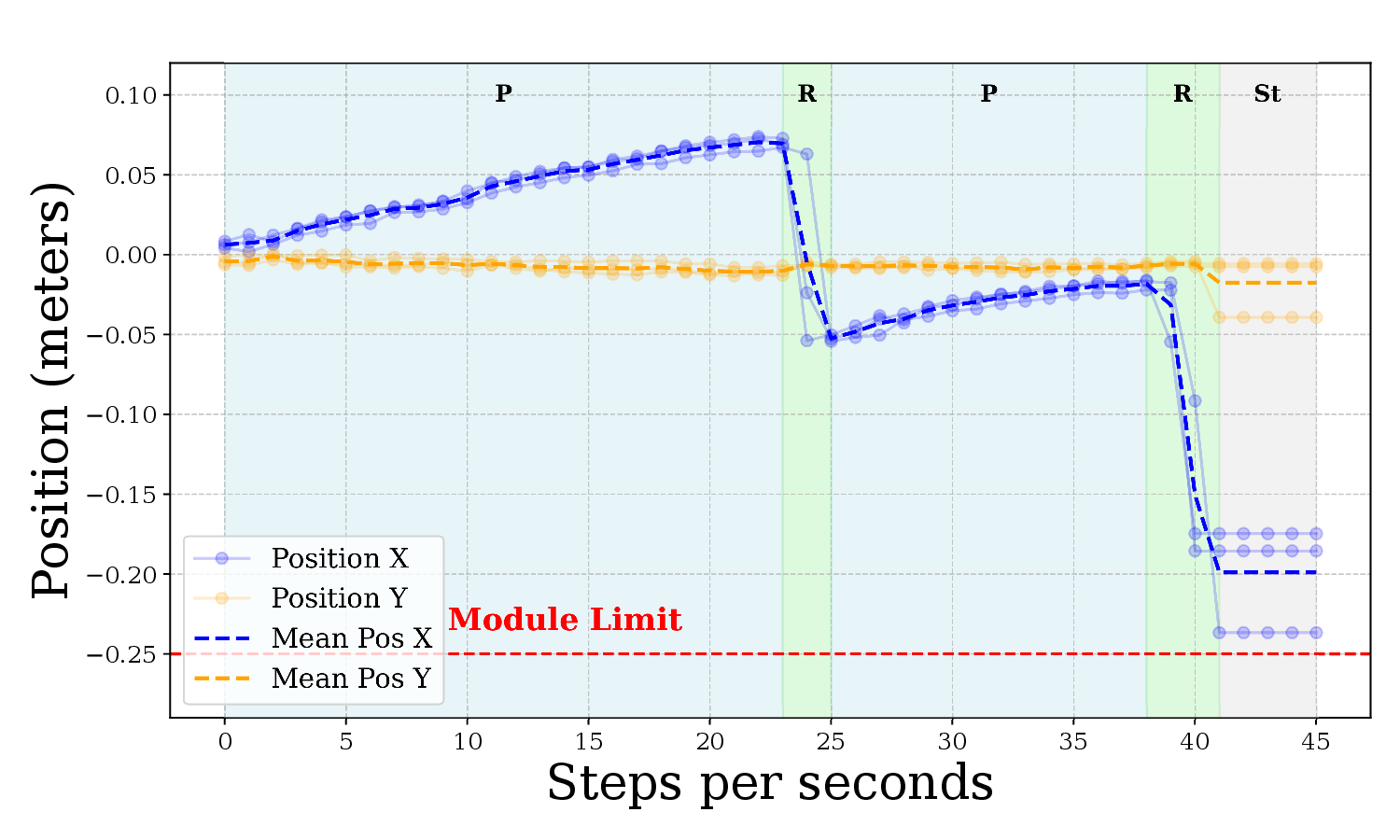} 
        \caption{Cube dynamics}
        \label{fig:r_edge_cube}
    \end{subfigure}
    \hfill
    \begin{subfigure}[b]{0.32\textwidth}
        \centering
        \includegraphics[width=\textwidth]{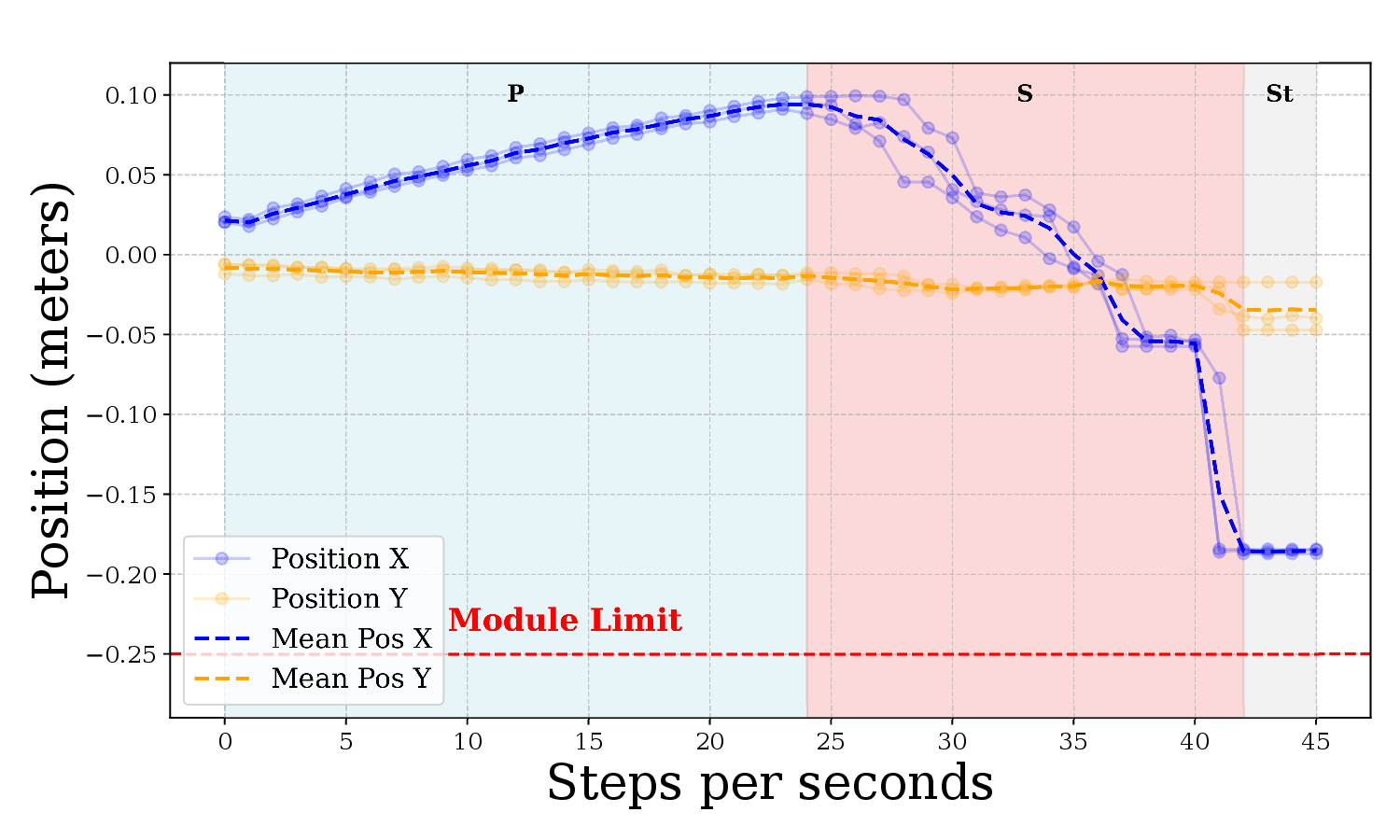} 
        \caption{Disk dynamics}
        \label{fig:r_edge_disk}
    \end{subfigure}
    
    \caption{Object movement toward the edge, with shaded regions for rolling (R, green), sliding (S, red), pulling (P, blue), and stationary (St, gray) behaviors. Red dashed line marks module limit}
    \label{fig:rate_edge}
\end{figure*}

\begin{figure*}[ht]
    % \centering
    % First row with three columns
    \begin{subfigure}[b]{0.32\textwidth}
        \centering
        \includegraphics[width=\textwidth]{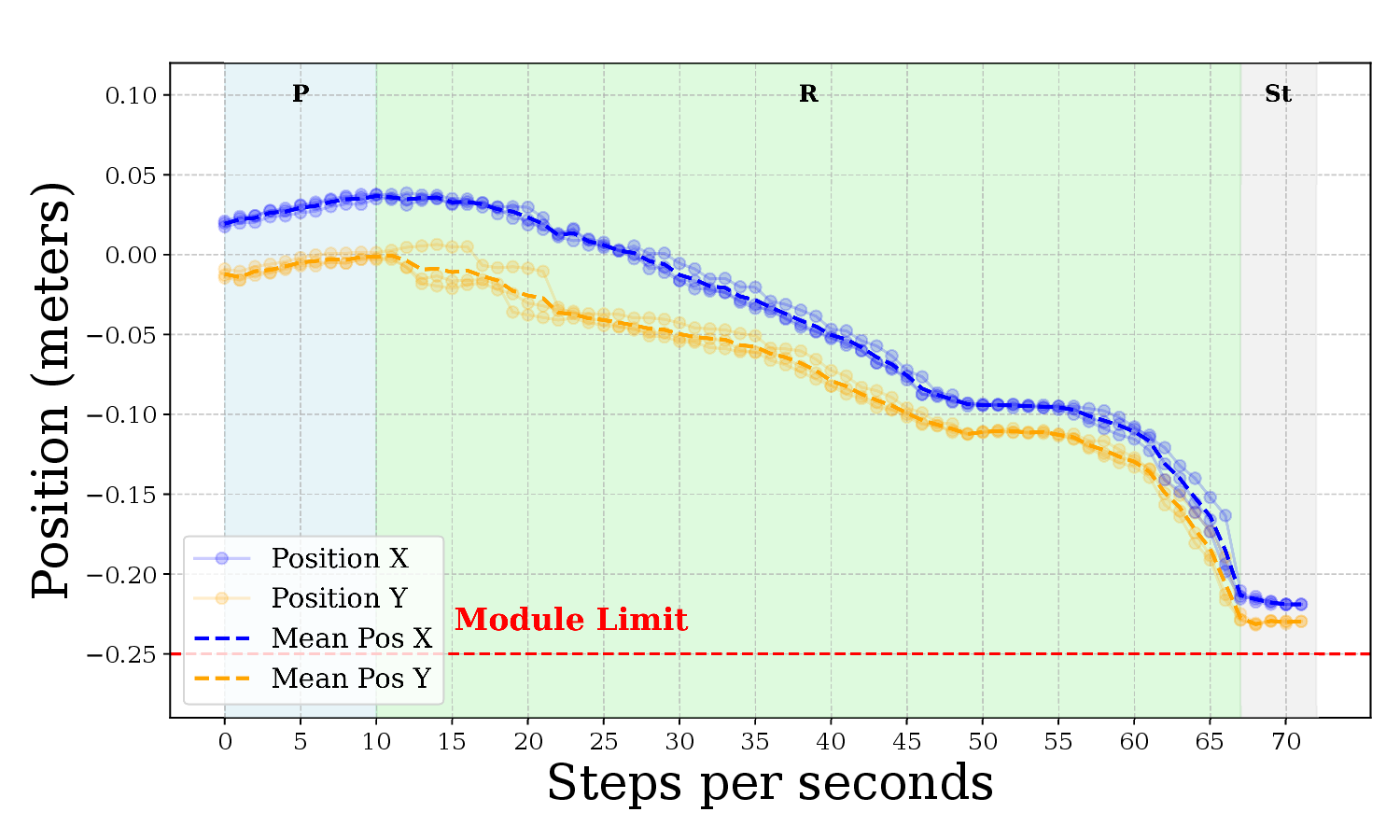} 
        \caption{Sphere dynamics}
        \label{fig:r_dia_sphere}
    \end{subfigure}
    \hfill
    \begin{subfigure}[b]{0.32\textwidth}
        \centering
        \includegraphics[width=\textwidth]{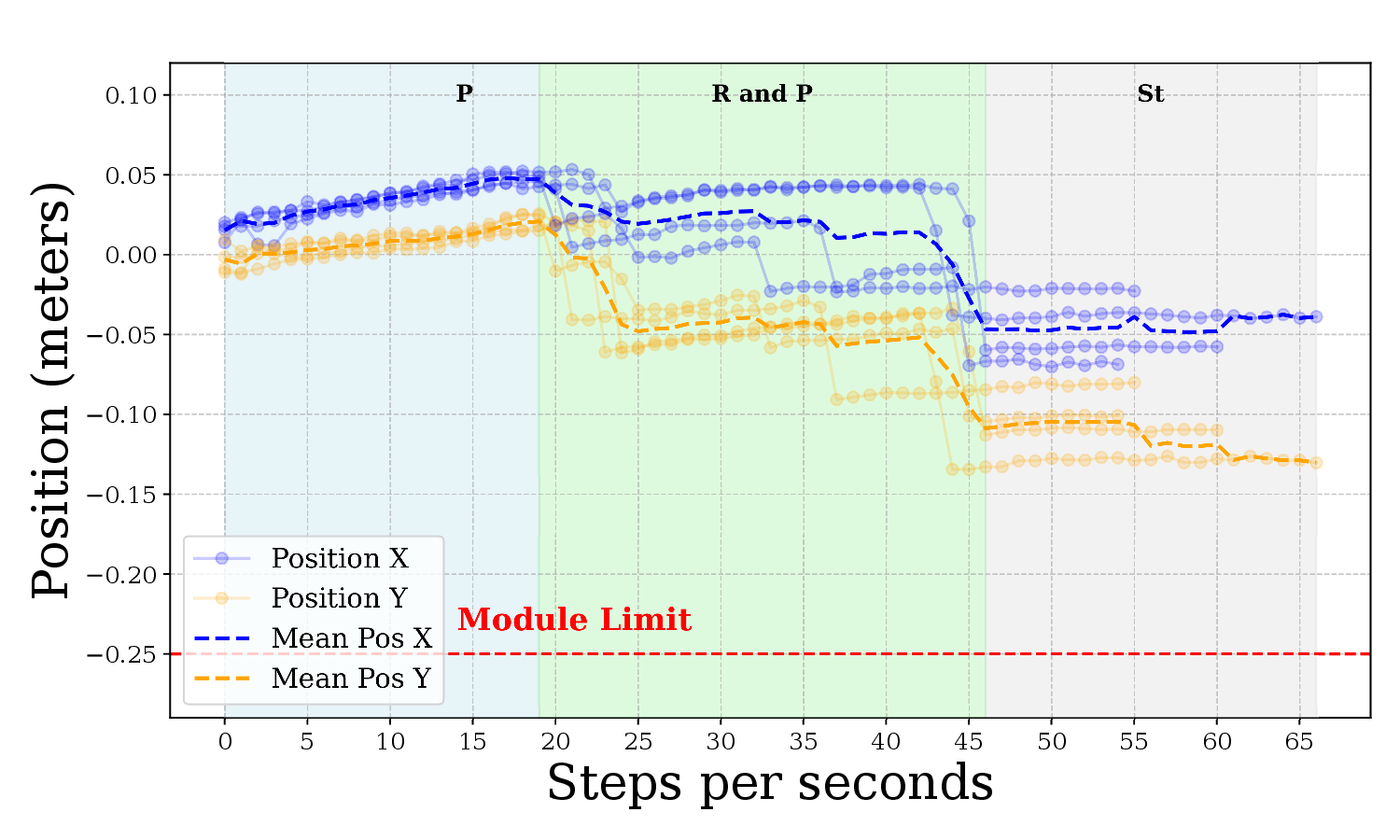} 
        \caption{Cube dynamics}
        \label{fig:r_dia_disk}
    \end{subfigure}
    \hfill
    \begin{subfigure}[b]{0.32\textwidth}
        \centering
        \includegraphics[width=\textwidth]{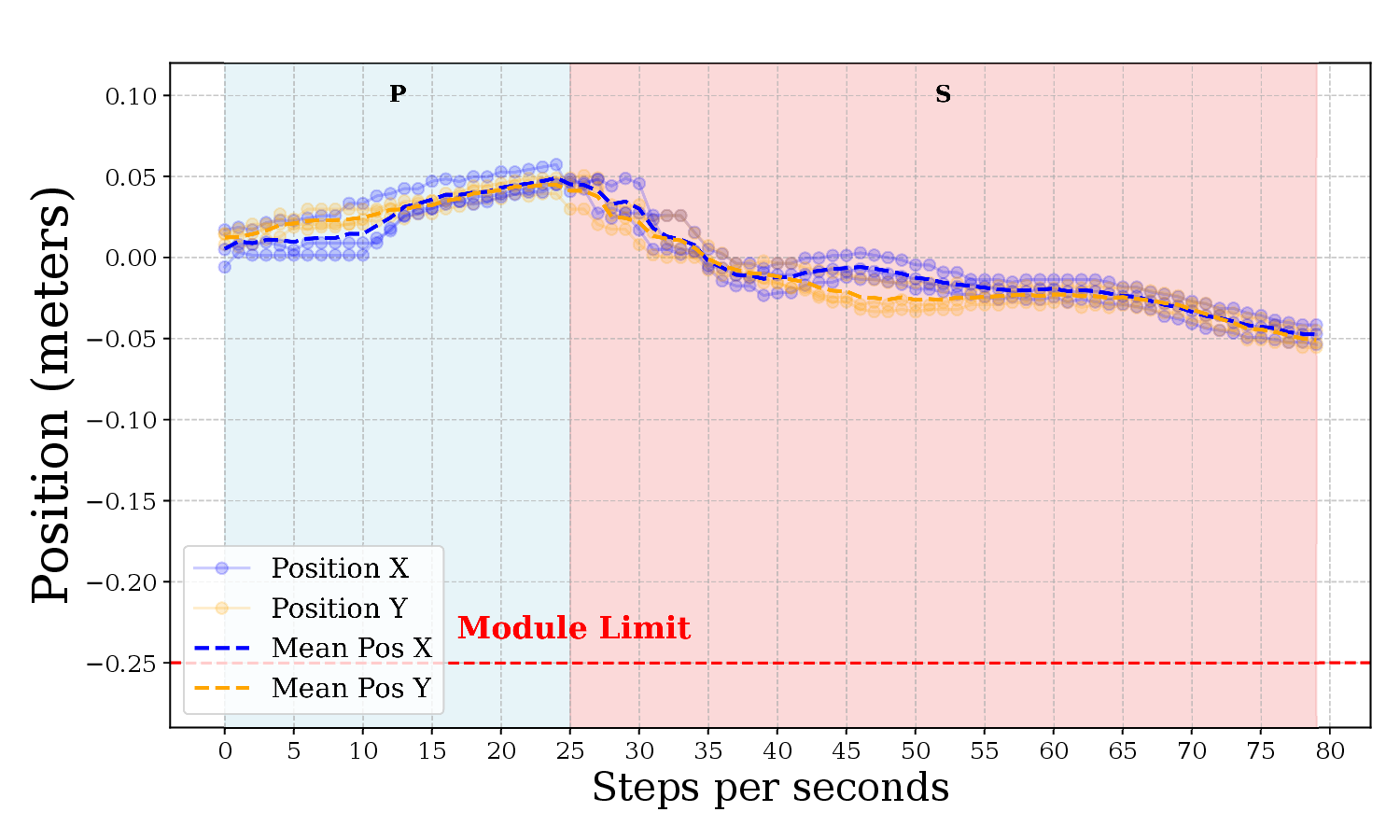} 
        \caption{Disk dynamics}
        \label{fig:r_dia_cube}
    \end{subfigure}
    
    \caption{Object movement toward the diagonal, with shaded regions for rolling (R, green), sliding (S, red), pulling (P, blue), and stationary (St, gray) behaviors. Red dashed line marks module limit}
    \label{fig:rate_diagonal}
\end{figure*}

A primary method for manipulating an object on soft fabric involves inducing rolling, sliding, or pulling motions (Object poison does not change with reference to the fabric). The tendency of an object to roll or slide depends largely on its shape. For example, a smooth, circular object is more likely to exhibit rolling motion, while a flat object with a low center of gravity will tend to slide rather than roll. The friction between the object and the soft fabric also influences these behaviors. Additionally, the motion can vary depending on whether the object is moving towards the edge or along the diagonal of the surface.

To observe these behaviour dynamics, we tested three objects of different shapes and weights: a sphere, a disk, and a cube. Each object was placed at the centre of the soft surface and manipulated in two directions: (1) toward the edge of the surface and (2) toward the corner of the surface along the diagonal. The object properties are described in the table \ref{tab:objects}.

\begin{table}[h!]
\centering
\begin{tabular}{l l l l}
\Xhline{3\arrayrulewidth} % Thicker top line
\addlinespace[2pt] % Add space between top line and header
\textbf{Object} & \textbf{Weight (g)} & \textbf{Size (cm)} & \textbf{Fabrication Method} \\ 
\addlinespace[2pt]
\Xhline{1\arrayrulewidth} % Regular line between header and content
\addlinespace[2pt]
Sphere    & 12.5  & \diameter: 5.1           & Molding        \\ 
Cube      & 66    & 5 x 5 x 5               & 3D Printing    \\ 
Disk      & 3.8   & \diameter: 4, Thickness: 0.4 & 3D Printing    \\ 
Apple     & 150   & \diameter: 5--7.5        & Supermarket \\ 
Cylinder  & 25.7  & \diameter: 4.7, Height: 4   & Molding        \\ 
Egg       & 68.6  & \diameter: 4.4--6.2         & Supermarket      \\ 
\addlinespace[2pt] % Add space between the last row and bottom line
\Xhline{3\arrayrulewidth} % Thicker bottom line
\end{tabular}
\caption{Details of five objects: weight, size, and fabrication method.}
\label{tab:objects}
\end{table}

To move an object toward the edge, we kept two actuators on one edge stationary at zero height and incrementally increased the elevation of the two actuators on the opposite edge by 1 cm per step at a rate of 1 step per second. For diagonal movement toward the corner, we kept one actuator stationary at zero reference height, moved the actuator on the opposite diagonal corner by 1 cm per second per step, and incrementally raised the two neighboring actuators by 0.5 cm per second per step from the zero reference until they reached maximum height. The results of the object movement toward the edge are shown in Figure \ref{fig:rate_edge}, and the object movement toward the corner along the diagonal is shown in Figure \ref{fig:rate_diagonal}, presented as line plots of the X and Y positions of the objects and the mean position over three runs.

In Figure \ref{fig:rate_edge}, each subplot represents data for an object, with the most notable changes observed along the X-axis as the object moved toward the edge in a straight path, while the position along the Y-axis remained relatively constant, demonstrating efficient movement along straight lines. In Figure \ref{fig:rate_diagonal}, both X-axis and Y-axis values are observed. Depending on the shape of the object, manipulation exhibited different behaviors—rolling (R, light green), sliding (S, light red), pulling (P, light blue), or stationary (St, light grey)—represented by different shaded regions. The red dashed line indicates the module boundary. Spherical objects tended to roll and fall off the surface with small elevation changes, while non-spherical objects like disks managed to slide without leaving the surface. These findings could be valuable for multi-module setups, where objects are transferred between modules, and they suggest the need for a more efficient controller.

Motion toward the corner along the diagonal was smoothly achieved by the spherical object, which replicated the same behavior consistently across multiple runs. In contrast, while the cube and disk exhibited similar behavior in each run, they did not travel as far along the diagonal. This indicates the increased complexity of object control near the corners of the modules. These results were obtained in an experiment using slow, quasi-stationary actuator motion. Greater movement may be achieved by faster and more dynamic motions, as will be shown in the next experiment.

\subsection{Object Behaviour}

\begin{figure*}[ht]
    % \centering
    % First row with three columns
    \begin{subfigure}[b]{0.32\textwidth}
        \centering
        \includegraphics[width=\textwidth]{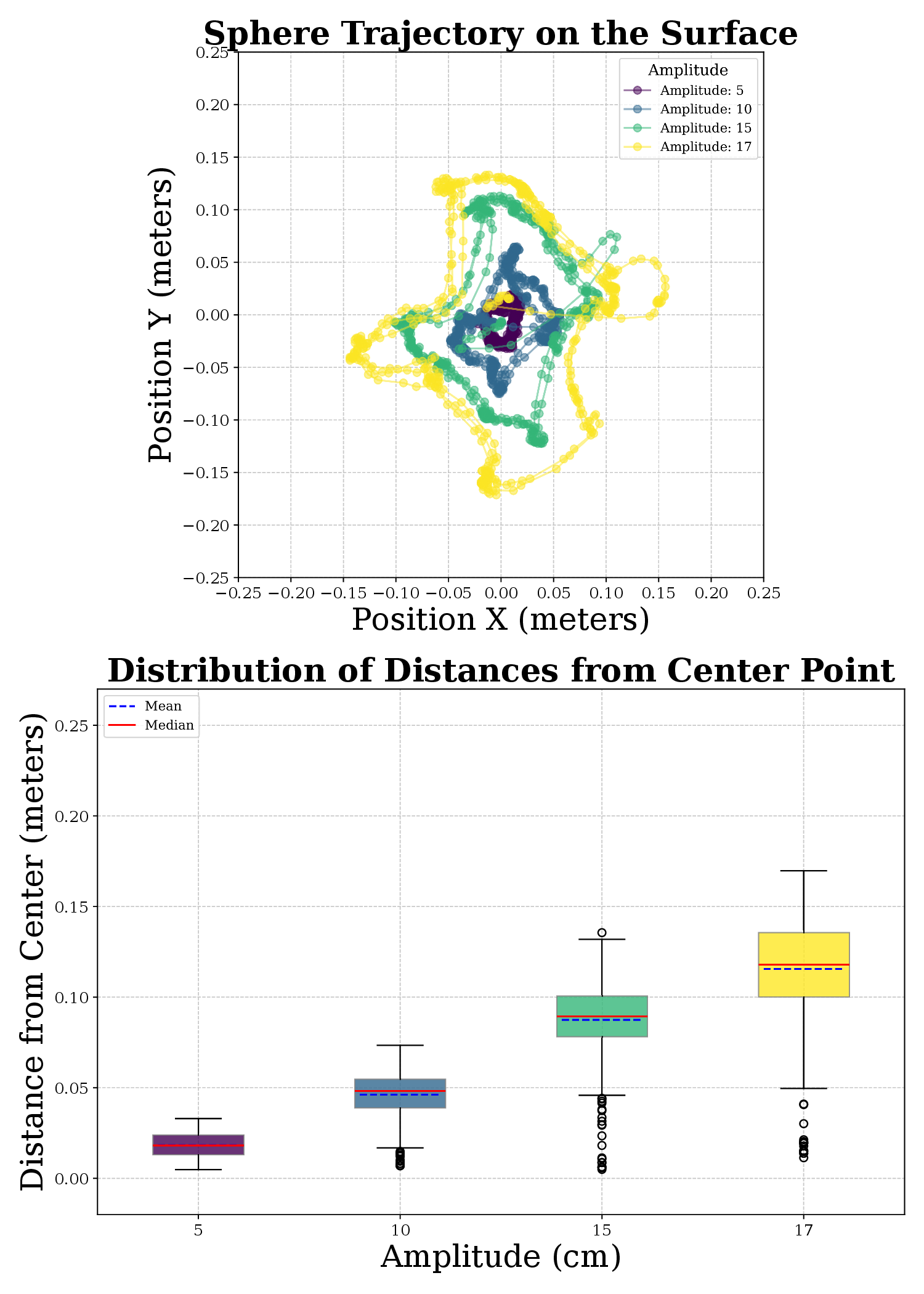} 
        \caption{Sphere}
        \label{fig:b_sphere}
    \end{subfigure}
    \hfill
    \begin{subfigure}[b]{0.32\textwidth}
        \centering
        \includegraphics[width=\textwidth]{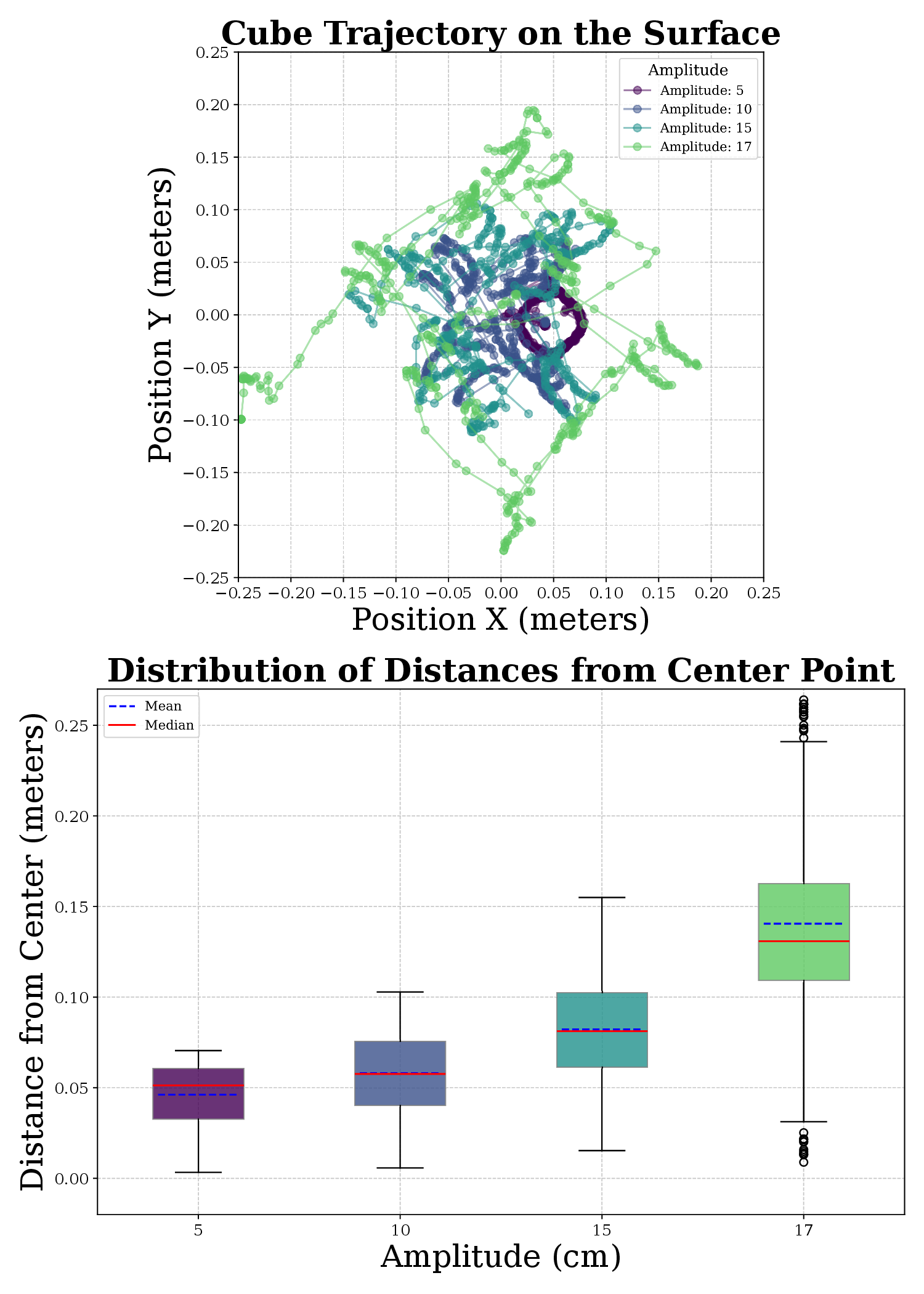} 
        \caption{Cube}
        \label{fig:b_cube}
    \end{subfigure}
    \hfill
    \begin{subfigure}[b]{0.32\textwidth}
        \centering
        \includegraphics[width=\textwidth]{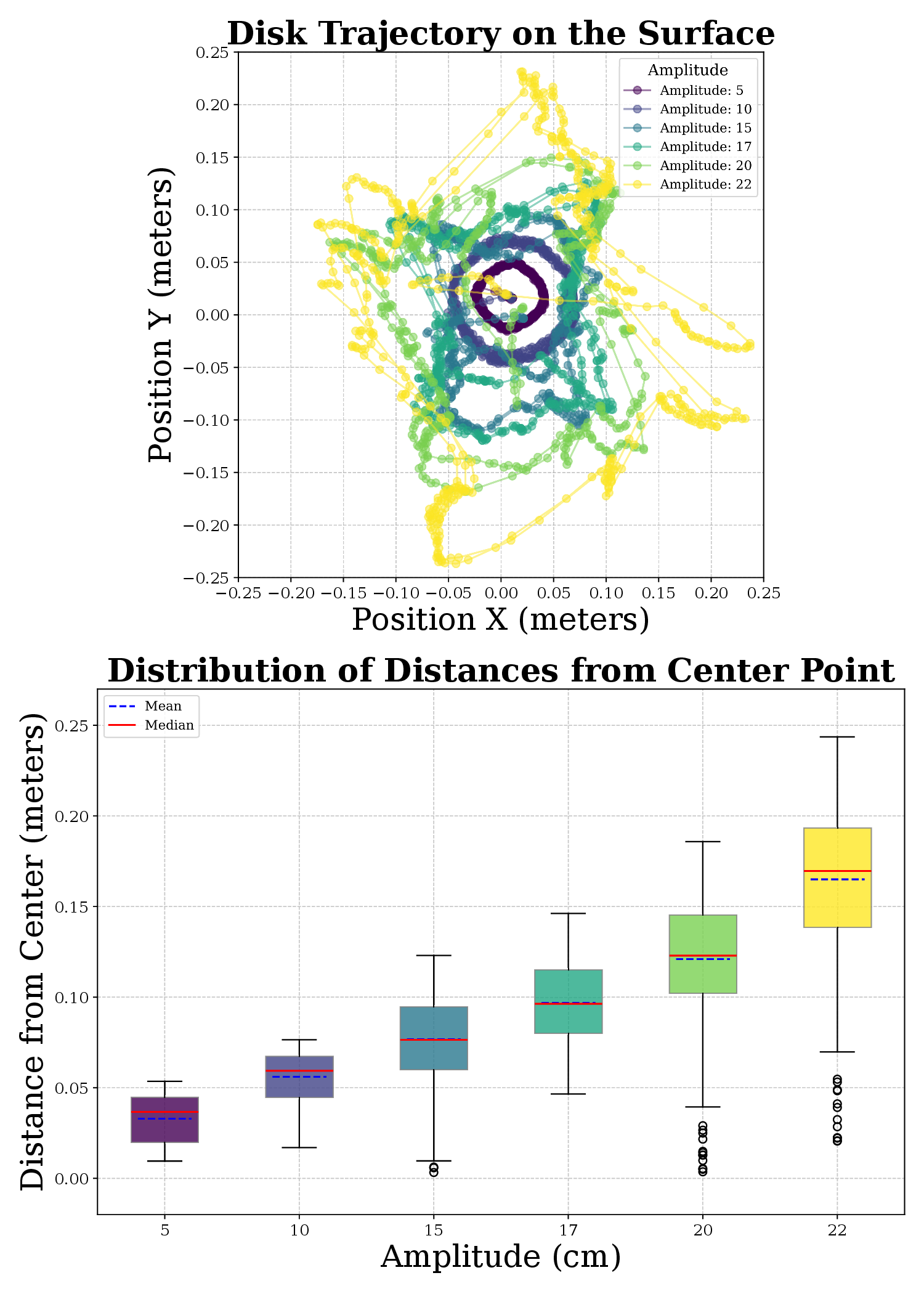} 
        \caption{Disk}
        \label{fig:b_disk}
    \end{subfigure}
    \hfill
    \begin{subfigure}[b]{0.32\textwidth}
        \centering
        \includegraphics[width=\textwidth]{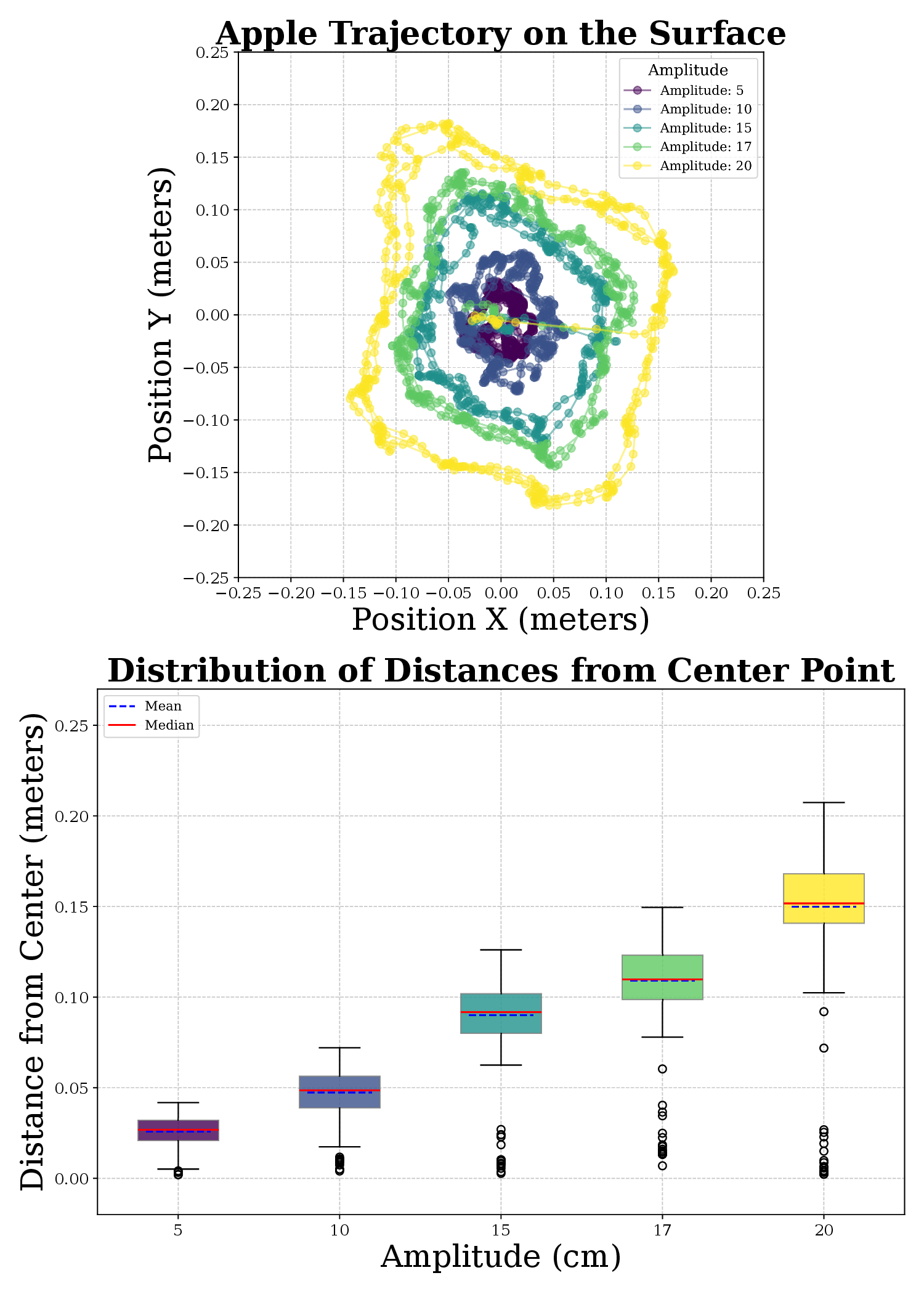} 
        \caption{Apple}
        \label{fig:b_apple}
    \end{subfigure}
    \hfill
    \begin{subfigure}[b]{0.32\textwidth}
        \centering
        \includegraphics[width=\textwidth]{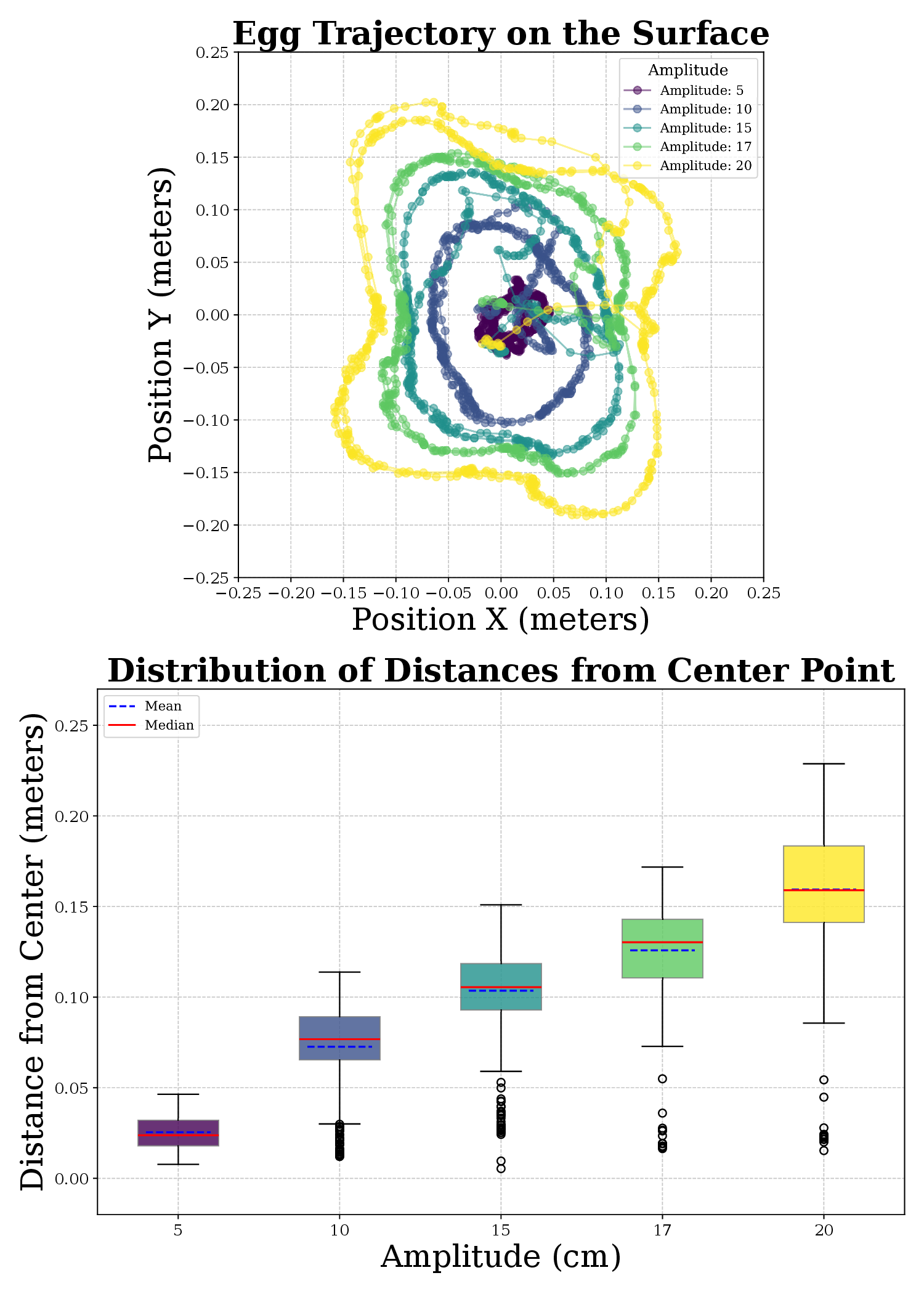} 
        \caption{Egg}
        \label{fig:b_egg}
    \end{subfigure}
    \hfill
    \begin{subfigure}[b]{0.32\textwidth}
        \centering
        \includegraphics[width=\textwidth]{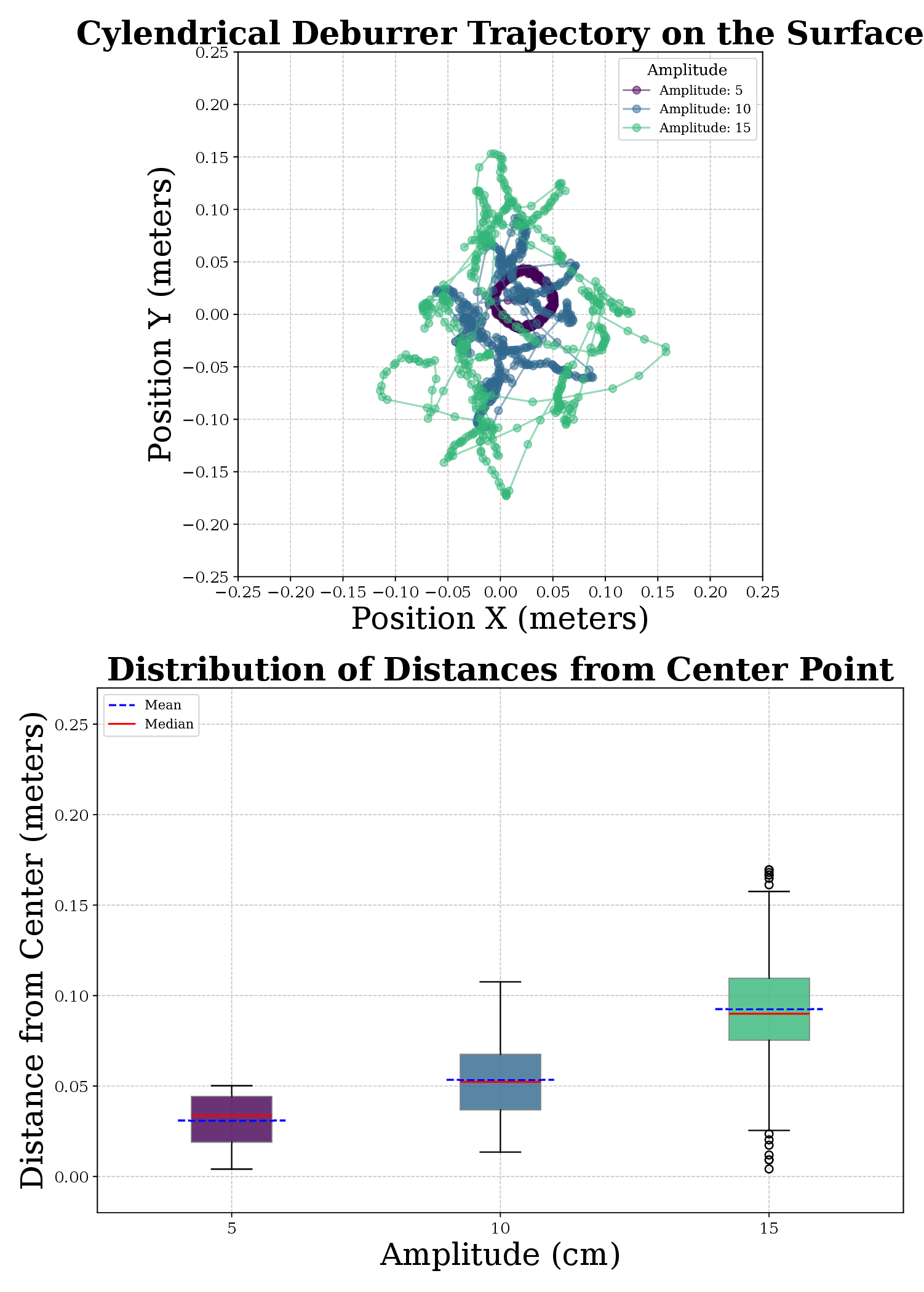} 
        \caption{Cylindrical Deburrer}
        \label{fig:b_random}
    \end{subfigure}
    
    \caption{Circular motion of various objects on the surface. The top graph shows object trajectories at different amplitudes, while the bottom graph presents a box plot of the distribution of each object's distance from the fabric centre.}
    \label{fig:behaviour}
\end{figure*}

The soft fabric layer is connected to the actuators at the corners, creating an uneven surface due to sagging in the center and stretching or folding at the edges from the fabric's weight. This folding and stretching increase further when a heavy object is placed on top for manipulation, making it difficult to manipulate objects, particularly near the corners and edges. To assess the extent of this limitation, we moved objects along a circular trajectory on the fabric by varying the actuators in a sinusoidal wave, with each actuator out of phase by $\pi/4$ relative to the others and moving with different amplitudes. The smallest amplitude was 5 cm, increasing in 5 cm increments until the object fell off the fabric. Each run lasted 1 minute, and object trajectories were recorded.

The same three objects as before—a cube, a sphere, and a disk—were used for the experiments. To evaluate the system's effectiveness with real-world objects, an apple, an egg, and a cylindrical-shaped deburrer were also tested (see Table \ref{tab:objects}). The deburrer, while cylindrical in shape, is mostly hollow with a conical structure inside, resulting in a different center of mass compared to a solid cylinder. It has a textured edge surface but no top or bottom surfaces. Depending on its position, it may roll, flip, or slide. This unique shape presents a particular challenge for manipulation using traditional RMS.

The results are shown in Figure \ref{fig:behaviour}. At low amplitudes, all objects followed a circular path because they did not slide, roll, or flip but were instead pulled by the fabric. This is especially noticeable in the disk, as seen in Figure \ref{fig:b_disk} for amplitudes of 5 and 10 cm. As the amplitude increased, the objects began sliding or rolling, and the circular path became more rectangular due to the stretching of the fabric near the corners. The plot also shows the distribution of distance with respect to amplitude as a box plot, illustrating that greater amplitude results in greater distance traveled. Interestingly, the apple (Figure \ref{fig:b_apple})), which closely resembles a sphere, performed better than the actual sphere (Figure \ref{fig:b_sphere}). This improved performance is attributed to the weight difference between the apple and the sphere, indicating that an object’s weight plays a significant role in manipulating the soft surface, as it helps to overcome folds or irregularities in the fabric.

It has been observed that when a heavy spherical object rolls, its path initially resembles a circle but distorts into a rectangular shape due to the fabric's stretching effect, as seen with the apple. For lightweight spherical objects, such as the lightweight sphere, fabric folding dominates the path behavior. In the case of the cube, although it rolls, its trajectory at high amplitudes resembles a rectangular path but is more irregular. At amplitudes greater than 22 cm, all objects fell off the fabric. The cylindrical deburrer remained on the surface up to 15 cm, the sphere and cube up to 17 cm, and the apple and egg up to 20 cm. The disk stayed on the surface up to 22 cm, likely due to its flat surface, light weight, and low center of mass.

\section{Target reaching proof of concept}
\subsection{Pretraining in simulation}
\begin{figure}[h]
    \centering
    \includegraphics[width=\linewidth]{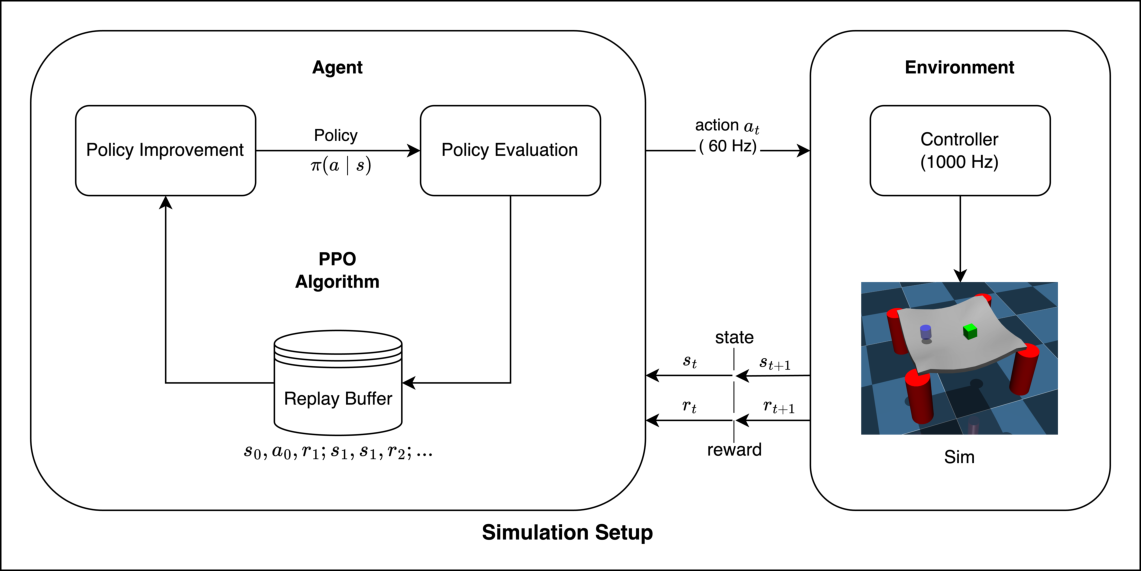}
    \caption{Simulation setup for training the policy network with agent and environment in the MuJoCo simulator.}
    \label{fig:simulation_setup}
\end{figure}

Controlling an object on a soft surface presents significant challenges due to the non-linear behavior of the surface, requiring the controller to account for these complexities. All manipulation occurs through indirect waves generated by the actuators on the soft surface. To develop a robust controller for this system, we employ reinforcement learning (RL), specifically training a Proximal Policy Optimization (PPO) \cite{schulman2017proximal} policy in a MuJoCo simulation environment (see Figure \ref{fig:simulation_setup}).

The system uses PPO to provide position values for the actuators ($a = {a_1, a_2, a_3, a_4}$), which control the actuators’ movements learned through simulation. In the simulation, the setup consists of four linear actuators, each capable of 0.5 meters of movement along the z-axis. These actuators (shown in red) are positioned at the corners of a $1 \times 1$-meter square, connected by a flexible soft layer (shown in gray) resting on top.

The primary objective is to manipulate an object (shown in green, with position $p_{\text{object}} = (p_x, p_y)$)—in this case, a sphere—to the desired target location (blue region) from any arbitrary starting position on the soft surface. To achieve this, PPO is trained using a reward function composed of four sub-rewards: (1) distance from the target, (2) velocity vector, (3) negative reward for falling off the surface, and (4) positive reward for reaching the target position.

\begin{figure}[h]
    \centering
    \includegraphics[width=\linewidth]{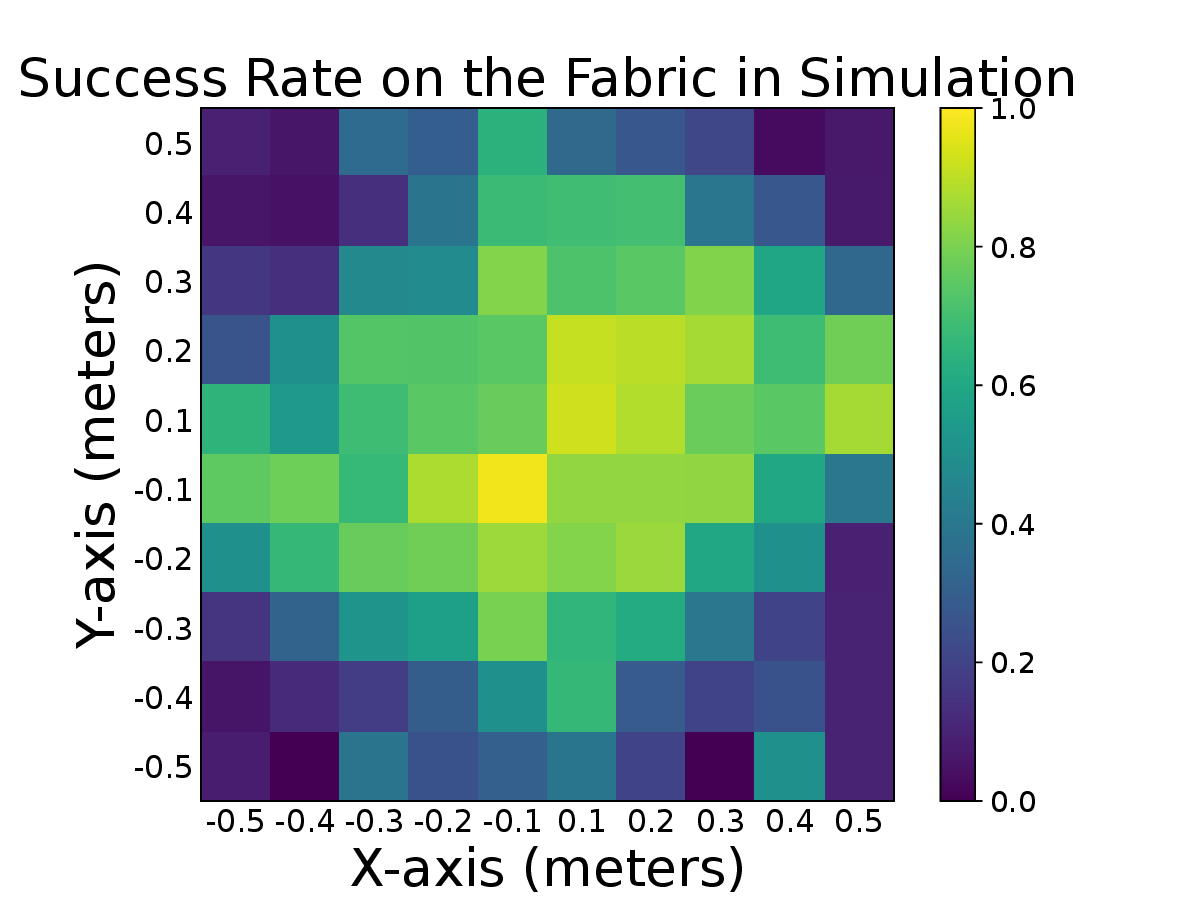} 
    \caption{Success rate of reaching target poison across the surface in simulation, with brighter regions indicating higher manipulation success}
    \label{fig:sim_sucess}
\end{figure}

The policy network is trained for 300K steps in the simulation. A sphere is used for policy training. In the simulation, we also evaluated the trained agent's success rate. To do this, the $1 \times 1$-meter region of interest was divided into 100 smaller sub-regions, each measuring $0.1 \times 0.1$-meters. A random target was assigned within each subregion, and the object was placed at a random starting position on the surface. The goal was to reach the target position without the object falling off the surface. This process was repeated 10 times for each subregion, and the mean success rate was calculated. The overall results are shown in Figure \ref{fig:sim_sucess}.

The trained policy network was highly effective in manipulating spheres within the simulation, particularly near the centre of the soft surface, achieving a 100\% success rate in reaching the target. However, the system struggled with edge cases due to high nonlinearity and catenary effect around the corners. 

\subsection{Demonstration and sim-to-real}

We perform a preliminary test of the policy controller on the hardware by transferring the control from the simulation to the physical setup with zero-shot learning to control an object on the hardware setup. The policy operates at a 60 Hz control frequency in the simulation, while the hardware operates at a reduced control frequency of 10 Hz due to the bottleneck caused by the camera’s position capture rate. The policies trained on spheres in the simulation were used to control the sphere, cube, and cylindrical deburrer on the hardware. We were able to move the object to the target location for some objects (Cube and Cylindrical Deburrer). However, the trained policy lacks robustness in the hardware experiments due to the sim-to-real gap and the difference in control frequencies. Figure 8 illustrates these results.
Cube and Cylindrical Deburrer achieve the target position, but the actions performed are not efficient. For the sphere, the lack of control frequency resulted in the object falling out of the surface as seen in Figure \ref{fig:r_dia_sphere}. Further work is needed to increase the robustness of RL control. 

\begin{figure*}[ht]
    % \centering
    % First row with three columns
    \begin{subfigure}[b]{0.32\textwidth}
        \centering
        \includegraphics[width=\textwidth]{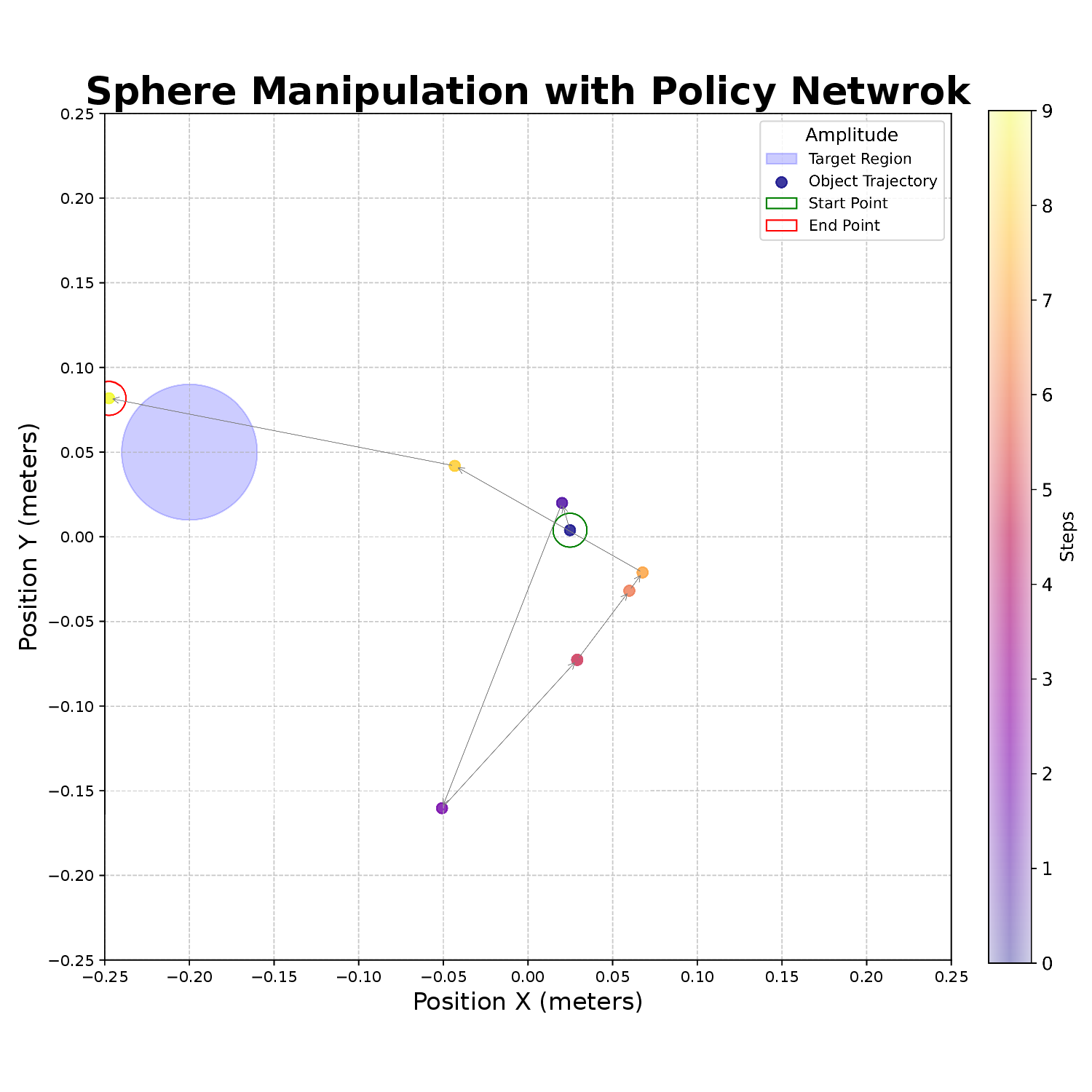} % Replace with your image path
        \caption{Sphere}
        \label{fig:r_dia_sphere}
    \end{subfigure}
    \hfill
    \begin{subfigure}[b]{0.32\textwidth}
        \centering
        \includegraphics[width=\textwidth]{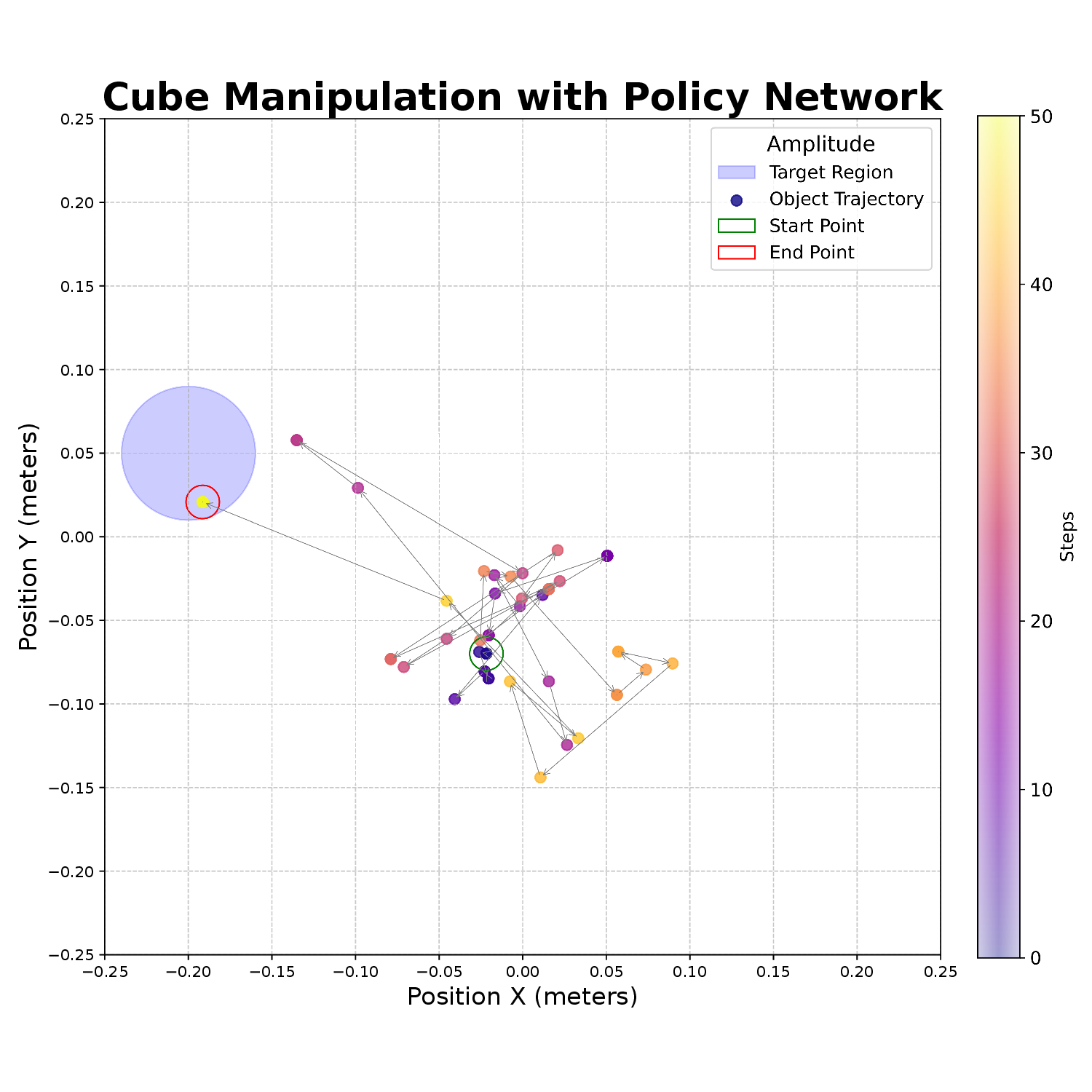} % Replace with your image path
        \caption{Cube}
        \label{fig:r_dia_disk}
    \end{subfigure}
    \hfill
    \begin{subfigure}[b]{0.32\textwidth}
        \centering
        \includegraphics[width=\textwidth]{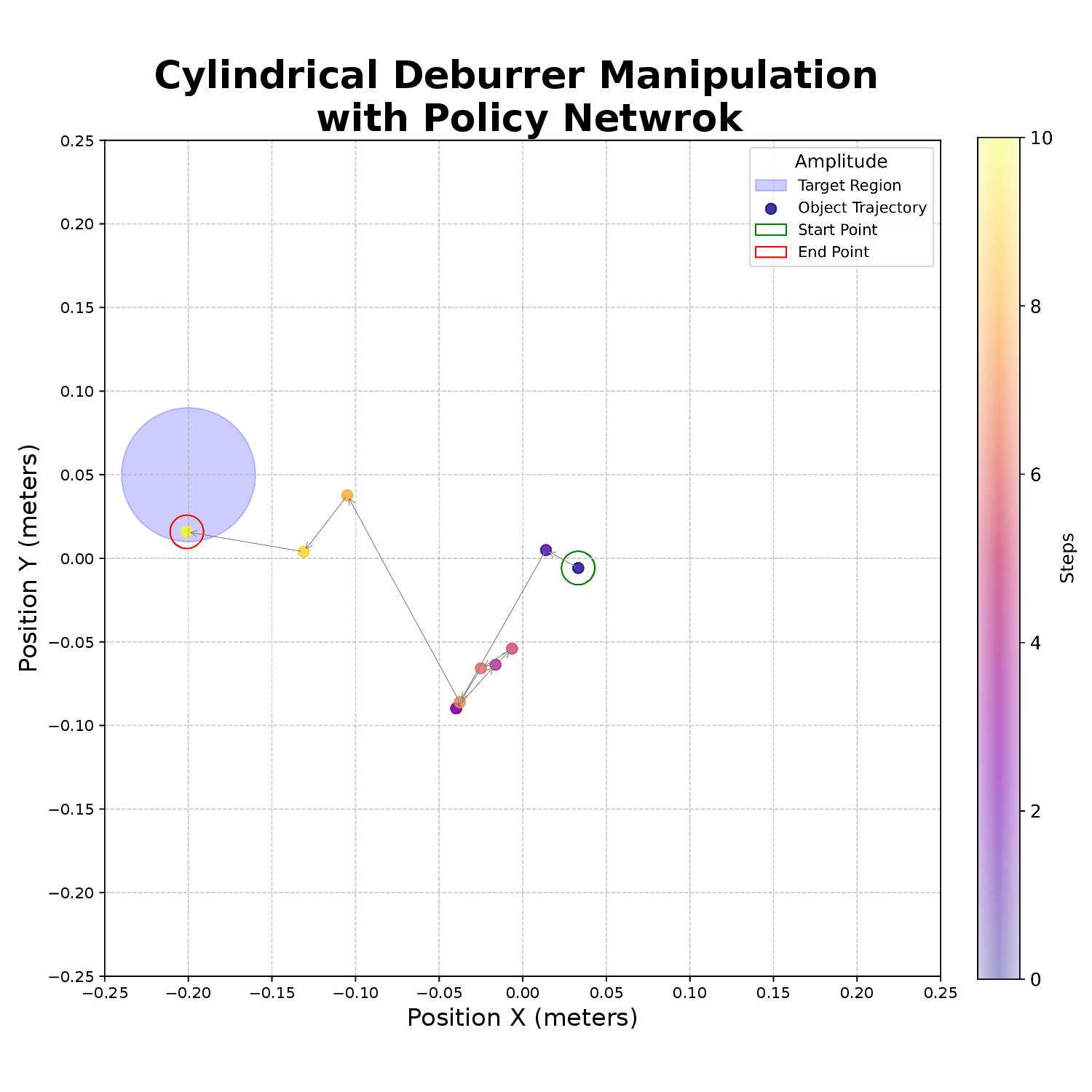} % Replace with your image path
        \caption{Cylindrical Deburrer}
        \label{fig:r_dia_cube}
    \end{subfigure}
    
    \caption{Object manipulation in a physical system using a policy network. Green and red circles denote the object’s start
and end positions, respectively, with blue regions as target areas. Scatter points represent observed object positions, where
yellow indicates more recent observations. The line plot below shows action values based on these observations.}
    \label{fig:PN_target}
\end{figure*}
\section{Discussion}
The use of soft surfaces can significantly improve the handling of fragile objects, as the fabric applies minimal force and deforms in worst-case scenarios. Most manipulation occurs through sliding, rolling, and pulling motions, indirectly caused by the actuators, and soft surfaces can also cover large areas effectively.

Previous work on soft manipulation surfaces, such as in \cite{johnson2023multifunctional, jang2024dynamically, festo}, is limited, often restricted to spherical objects manipulated by rolling due to limited actuator reach and strength. In contrast, our system can handle objects of various shapes, sizes, and weights (see Table \ref{tab:objects}).

This study focuses on a single module with dimensions of $0.5 \times 0.5$-meters and a $0.6 \times 0.6$-meter cloth for manipulation, though both the module and fabric size can be adjusted as needed. To cover larger distances, actuators could be spaced further apart, but this would also require increasing each actuator’s range to maintain the desired slope. Additionally, the size of the soft surface plays a role: smaller cloths would need to stretch, while larger cloths would cause objects to “hang” excessively, limiting manipulation feasibility. The soft layer material can also be replaced to modify friction, softness, or stretchability, depending on the object type. Using multiple modules allows for better control and simultaneous manipulation of multiple objects, as modules can vary in size and fabric properties to support diverse manipulation behaviors. These modules can be replicated and combined to scale the system for larger tasks.

However, using soft layers for object handling presents the challenge of achieving robust control due to the fabric’s non-linear behavior. While reinforcement learning (RL) provides more robust control compared to classical strategies, training these models is complex. The MuJoCo simulator approximates cloth and soft body behavior using composite objects, but this simulation often suffers from a reality gap \cite{blanco2024benchmarking}. Imperfect replication of fabric folds leads to a sim-to-real gap, degrading real-world performance. Real-world RL, where policies are directly trained on hardware, could address this gap but involves high costs and lengthy training times. Experimental findings of system limitations and dependencies of manipulation behavior on object properties on real hardware, as discussed in section III Experiments/Results, could be used to improve the RL policy's robustness further. 

When the soft layer is supported by four actuators, a catenary effect occurs, making manipulation easier near the center of the surface. However, real fabric has wrinkles and an uneven texture, causing local perturbations that affect lighter objects more significantly. In contrast, heavier objects (relative to the weight of the fabric) tend to flatten these irregularities by deforming the surface. Reducing fabric softness could also help minimize unwanted folds.

Our system effectively manipulates a single object on one module (four actuators), though handling multiple objects on the same surface results in correlated behaviors. Parallel manipulation of multiple objects can be achieved by using multiple modules. Multi-agent reinforcement learning could be well-suited for controlling such a system, enabling decentralized control, improved fault tolerance, and increased robustness. This modular approach would allow for large-scale coverage, with decentralized modules of varying sizes and surfaces working together to handle diverse objects and manipulation properties, further enhancing system robustness and fault tolerance.

\section{Conclusion}
We propose a new system that uses a soft layer as the primary means of manipulation. In this system, objects are manipulated on the soft surface using actuators. These actuators can be spaced farther apart, allowing the system to cover larger areas with fewer actuator densities compared to existing approaches, and the area of 0.5x0.5 meters is customizable. This reduces the system’s density and makes it easier to maintain. Additionally, it resolves the issue of object size constraints, enabling the manipulation of objects much smaller than the actuators or the gaps between them. The soft layer is also ideal for handling fragile objects, and depending on the object’s characteristics, the soft layer can be customized with specific physical properties such as friction, softness, or elasticity. 

Our experiments on real hardware demonstrated that this system can effectively manipulate objects of various shapes, sizes, and weights. We also showed that it can handle delicate objects, such as eggs and apples. To account for the non-linear behavior of the soft fabric, we employed reinforcement learning (RL) to control object manipulation in simulations, achieving up to a 100\% success rate in certain regions. Future work will explore more robust RL control strategies, decentralized controllers, and methods for bridging the reality gap over larger areas with multiple actuators to further enhance the system’s reliability and flexibility.

\bibliographystyle{unsrt}
\bibliography{ref}
\vspace{12pt}
\end{document}